\documentclass[sigconf]{acmart} %manuscript,sigconf

\usepackage{times}
\usepackage{url}
\usepackage{enumerate}
\usepackage{amsthm}
\usepackage{amsfonts}
\usepackage{diagbox}
\usepackage{multirow}
\usepackage{caption}
\usepackage{tabularx}

\usepackage[vlined,ruled,linesnumbered]{algorithm2e}
\usepackage{subfigure}
\usepackage{color}

\AtBeginDocument{%
  \providecommand\BibTeX{{%
    \normalfont B\kern-0.5em{\scshape i\kern-0.25em b}\kern-0.8em\TeX}}}

\copyrightyear{2020} 
\acmYear{2020} 
\setcopyright{acmlicensed}\acmConference[RecSys '20]{Fourteenth ACM Conference on Recommender Systems}{September 22--26, 2020}{Virtual Event, Brazil}
\acmBooktitle{Fourteenth ACM Conference on Recommender Systems (RecSys '20), September 22--26, 2020, Virtual Event, Brazil}
\acmPrice{15.00}
\acmDOI{10.1145/3383313.3412233}
\acmISBN{978-1-4503-7583-2/20/09}

\begin{document}

\title{Learning to Collaborate in Multi-Module Recommendation via Multi-Agent Reinforcement Learning without Communication}

% \author

% \author{Xu He$^1$, Bo An$^1$, Yanghua Li$^2$, Haikai Chen$^2$, Rundong Wang$^1$, Xinrun Wang$^1$, Runsheng Yu$^1$, Xin Li$^2$, Zhirong Wang$^2$}
% \affiliation{\institution{$^1$Nanyang Technological University, $^2$Alibaba Group}}
% \email{{hexu0003, boan}@ntu.edu.sg, {yichen.lyh, haikai.chk}@taobao.com, {rundong001, xinrun.wang, runsheng.yu}@ntu.edu.sg}
% \email{xin.l@alibaba-inc.com, qingfeng@taobao.com}

\author{Xu He, Bo An}
\affiliation{\institution{Nanyang Technological University}}
\email{{hexu0003,boan}@ntu.edu.sg}

% \author{Bo An}
% \affiliation{\institution{Nanyang Technological University}}
% \email{boan@ntu.edu.sg}

\author{Yanghua Li, Haikai Chen}
\affiliation{\institution{Alibaba Group}}
\email{{yichen.lyh,haikai.chk}@taobao.com}

% \author{Haikai Chen}
% \affiliation{\institution{Alibaba Group}}
% \email{haikai.chk@taobao.com}

\author{Rundong Wang, Xinrun Wang, Runsheng Yu}
\affiliation{\institution{Nanyang Technological University}}
\email{{rundong001, xinrun.wang, runsheng.yu}@ntu.edu.sg}

% \author{Xinrun Wang}
% \affiliation{\institution{Nanyang Technological University}}
% \email{xinrun.wang@ntu.edu.sg}

% \author{Runsheng Yu}
% \affiliation{\institution{Nanyang Technological University}}
% \email{runsheng.yu@ntu.edu.sg}

\author{Xin Li, Zhirong Wang}
\affiliation{\institution{Alibaba Group}}
\email{xin.l@alibaba-inc.com, qingfeng@taobao.com}

% \author{Zhirong Wang}
% \affiliation{\institution{Alibaba Group}}
% \email{qingfeng@taobao.com}

\renewcommand{\shortauthors}{Xu He et al.}
\renewcommand{\shorttitle}{Learning to Collaborate in Multi-Module Recommendation via \\ Multi-Agent Reinforcement Learning without Communication}

\begin{abstract}
  With the rise of online e-commerce platforms, more and more customers prefer to shop online. To sell more products, online platforms introduce various modules to recommend items with different properties such as huge discounts. A web page often consists of different independent modules. The ranking policies of these modules are decided by different teams and optimized individually without cooperation, which might result in competition between modules. Thus, the global policy of the whole page could be sub-optimal. In this paper, we propose a novel multi-agent cooperative reinforcement learning approach with the restriction that different modules cannot communicate. Our contributions are three-fold. Firstly, inspired by a solution concept in game theory named correlated equilibrium, we design a signal network to promote cooperation of all modules by generating signals (vectors) for different modules. Secondly, an entropy-regularized version of the signal network is proposed to coordinate agents’ exploration of the optimal global policy. Furthermore, experiments based on real-world e-commerce data demonstrate that our algorithm obtains superior performance over baselines.
\end{abstract}
%   Various types of web pages are designed by these platforms to attract users. One of the most popular types is the page divided into multiple modules and each module recommends items with a specific property such as huge discounts in a limited time. 



\maketitle

\section{Introduction}
\begin{figure}[!t]
\setlength{\belowcaptionskip}{-0.3cm}
\setlength{\abovecaptionskip}{0cm}
  \centering
  \subfigure[Example 1]{
  \includegraphics[width=.18\textwidth]{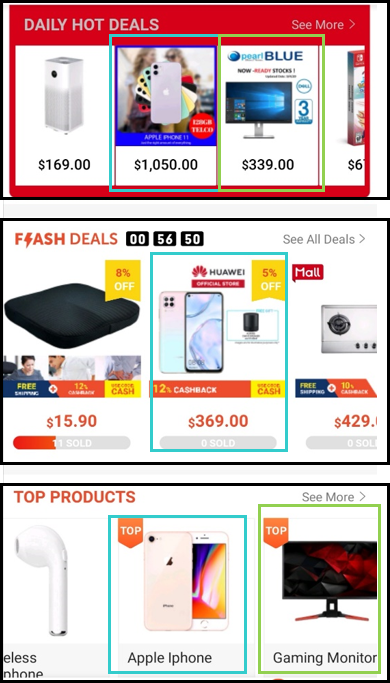}\label{module_1}
  }
  \subfigure[Example 2]{
  \includegraphics[width=.245\textwidth]{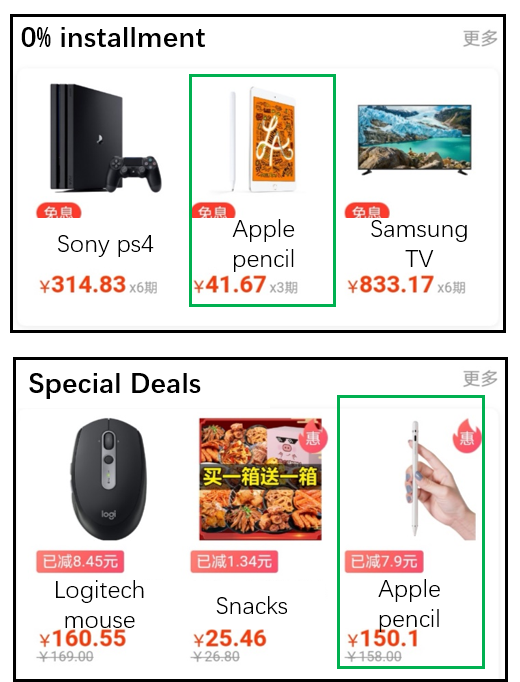}\label{module_2}
  }
  \caption{Two examples of the multi-module recommendation scenarios. The black boxes represent modules. Boxes in different colors mark similar items in different modules. In sub-figure \ref{module_1}, phones and monitors appear more than once. Meanwhile, apple pencils are recommended by two modules in sub-figure \ref{module_2}.}\label{module}
\end{figure}

% With the recent tremendous development in Reinforcement Learning (RL), more works adopt RL framework for recommendation due to two reasons. Firstly, RL algorithms can learn from real-time user-agent interaction continuously. Secondly, the optimal RL policy targets at maximizing long-term rewards rather than myopic rewards.
The web pages of many online e-commerce platforms consist of different modules. Each of the modules shows items with different properties. As an example, consider the web pages depicted in Fig. \ref{module}. The page on the left includes three modules: the daily hot deals, the flash sales, and the top products. There are two modules in the page on the right: the 0\% installment and the special deals. The candidate items of each module are selected according to predefined conditions. For instance, the top products module includes the best selling items in the period of the past few days. The items in the flash sales module and the special deals module offer special discounts provided by qualified shops, either daily or hourly. Because several modules are shown to users at the same time, the interaction between modules affects the users' experience.
% Thus, recommendation policies should be optimized considering the properties of candidate items from different modules. For example, 0\% installment module could be prone to recommend items with high prices and special deals module prefers items with huge discounts.

However, different teams are usually in charge of ranking strategies of different modules. Due to the lack of cooperation between the teams, the whole page suffers from competition between different modules. As a consequence, the users might find the same product or category in multiple modules, which wastes the limited space on the page. For example, the phones appear in all modules in Fig. \ref{module_1} and the apple pencil is recommended by two modules in Fig. \ref{module_2}. 
% However, the ranking strategies in different modules are usually decided by different teams to maximize the performance of each module. %and rarely consider the cooperation among modules
% Due to the lack of cooperation, each module is prone to compete with each other. This often leads to a sub-optimal global strategy for the whole web page. A direct consequence could be the repetition of items or categories on one web page. For example, phones appear at all modules in Fig. \ref{module_1} and apple pencil is recommended by two modules in Fig. \ref{module_2}. 
% It is wasting space to recommend the same item in different modules.

% However, most of the existing works \cite{zheng2018drn,zou2019reinforcement,pei2019value,zhao2018deep} focus on improving performance in a one-module environment. Thus, their methods lack cooperation when more than one agent (module) is involved.

To find the optimal global strategy, it is crucial to design a proper cooperation mechanism. 
Multi-agent reinforcement learning (RL) algorithms are proposed to solve the recommendation problems that involve sequential modules \cite{feng2018learning,DBLP:journals/corr/abs-1902-03987}. However, their approaches rely on an underlying communication mechanism. Each agent is hence required to send and receive messages during the execution. 
This might be a problem as ranking strategies of different modules are usually deployed by different teams in real-time and the modules cannot communicate with each other.
There are many examples of multi-agent RL algorithms in the literature which do not need communication. However, their performance suffers a lot from their inability to coordinate, as we illustrate in the experiments. 
In this paper, we propose a novel approach for the multi-module recommendation problem.
The first key contribution of this paper is a novel multi-agent cooperative reinforcement learning structure. The structure is inspired by a solution concept in game theory called correlated equilibrium \cite{aumann1974subjectivity} in which the predefined signals received by the agents guide their actions.
In our algorithm, we propose to use a signal network to maximize the global utility by taking the information of a user as input and sending signals to different modules. The signal network can act as a high-level leader coordinating the individual agents. All agents act solely on the basis of their signals, without any communication. 

The second key contribution is an entropy-regularized version of the signal network to coordinate agents’ exploration. Since the state and action spaces are huge, exploration remains essential in finding the optimal policy. We add the entropy terms to the loss function of the signal network to encourage exploration in view of the global performance. In contrast, the agents in the existing work \cite{iqbal2018actor} explore individually. To maximize the entropy term, the distributions of signals should be flat. In that case, the diverse signals encourage agents to explore more when the global policy converges to a sub-optimal solution.
% Some works  introduce entropy terms for the policy network to encourage exploration for each agent rather than the global policy. However, i

Third, we conduct extensive experiments on a real-world dataset from Taobao, one of the largest e-commerce companies in the world. Our proposed method outperforms other state-of-the-art cooperative multi-agent reinforcement learning algorithms. Moreover, we show the improvement caused by the entropy term in the ablation study. 

\section{Related Work}
% \subsection{RL applied in Recommender System}
We briefly review works that apply RL methods in recommender systems and introduce the concept of correlated equilibrium in this section.

Many deep reinforcement learning methods are used in the recommender system domain. 
The works focusing on the single-agent setting mainly consider three aspects: 1) the different kinds of rewards, 2) the structures of web pages and 3) the large space of actions.
DRN updates periodically after obtaining long term-reward such as return time \cite{zheng2018drn}. An algorithm is proposed to use two individual LSTM modules for items with short-term and long-term rewards respectively \cite{zou2019reinforcement}.
% The importance of long-term reward is learned by an individual structure \cite{zou2019reinforcement}. 
%More specifically, items with short-term and long-term rewards are embedded by two LSTM modules respectively.
%Reward shaping is a straightforward method to train a policy end to end by defining various reward functions. 
The diversity of recommended sets is added to the reward function \cite{liu2019diversity}. 
Transition probabilities from users' actions (such as click) to purchase are used as rewards \cite{pei2019value}. 
%A network is used to predict transition probabilities and the output is considered as rewards for actions, which is trained by historical data.
Modified MDPs for recommendation are proposed by redefining the structure of reward function and the transition function respectively \cite{ie2019reinforcement,hu2018reinforcement}. A method is proposed to improve profit by detecting fraud transactions \cite{zhao2018impression}. 
%Transition function is divided into three sub-functions that indicate the probabilities that users leave, purchase, or click.
Second, page structures including different types of content and positions of items are taken into consideration. A CNN-based approach is proposed to recommend items according to their positions on the web page \cite{zhao2018deep}. 
A hierarchical algorithm is proposed to aggregate topics, blog posts, and products on one web page \cite{takanobu2019aggregating}. 
Similarly, the Double-Rank Model is proposed to learn how to rank the display positions and with which documents to fill those positions \cite{oosterhuis2018ranking}.
%The positions of topics, posts and products are firstly decided by the high-level policy and low-level policies determine what to recommend.
The problem `when and where should advertising be added?' is addressed for web pages that contain advertising \cite{zhao2019deep}. 
%With a given recommended list, the rewards of additional advertising in different positions are learned.
% Since the positive reward is sparse, a work treats negative samples as additional information in training \cite{zhao2018recommendations}. 
%The main idea is to leverage two networks to process negative samples and positive samples respectively. 
Third, other works focusing on the large space of actions and states usually adopt clustering techniques to reduce the space \cite{choi2018reinforcement,sunehag2015deep}. To decide which items to recommend, the policy network outputs the feature of an ideal item and clustering methods are adopted to find neighbors of the ideal item in the candidate set of items.
% The importance sampling technique is proposed for off-policy training \cite{chen2019top}. 
However, these works do not consider recommendation problems that involve more than one agent and thus cannot be used to solve our problem.

The most similar works use multi-agent frameworks to promote cooperation between different pages.
Inspired by RL methods involving communication like \cite{wang2019learning}, a multi-agent RL framework is proposed where agents can communicate by messages \cite{feng2018learning}. A model-based RL algorithm is proposed by using neural networks to predict transition probability and shape reward \cite{DBLP:journals/corr/abs-1902-03987}. 
Differing from our setting, the agents in their works recommend items for different pages (e.g., the entrance page and the item detail page), and execute sequentially rather than simultaneously. It means that agents can send messages to others when users leave one page and enter another page. Moreover, the immediate reward is only related to one page (module). However, our problem considers cooperation between different modules on one page in which agents cannot communicate during execution, and the immediate rewards are determined by more than one module.

Correlated equilibrium is a solution concept in game theory, which is first discussed by \citeauthor{aumann1974subjectivity} \shortcite{aumann1974subjectivity}. The idea is that each player or agent chooses strategy according to their observation and a signal. The signal usually is a recommended strategy that assigns actions to all agents. If the expected payoff from playing the recommended strategy is no worse than playing any other strategy, it is called correlated equilibrium. An example is the traffic light, which suggests to each player whether to go or stop. Following its advice is the best response for everyone involved. A simple RL algorithm is proposed to find the correlated equilibrium in an existing work \cite{greenwald2003correlated}. In our work, we use a neural network to learn how to send signals inspired by this concept.

\section{Problem Statement and Formulation} \label{form}

\begin{figure*}[!t]
  \centering
  \includegraphics[width=.8\textwidth]{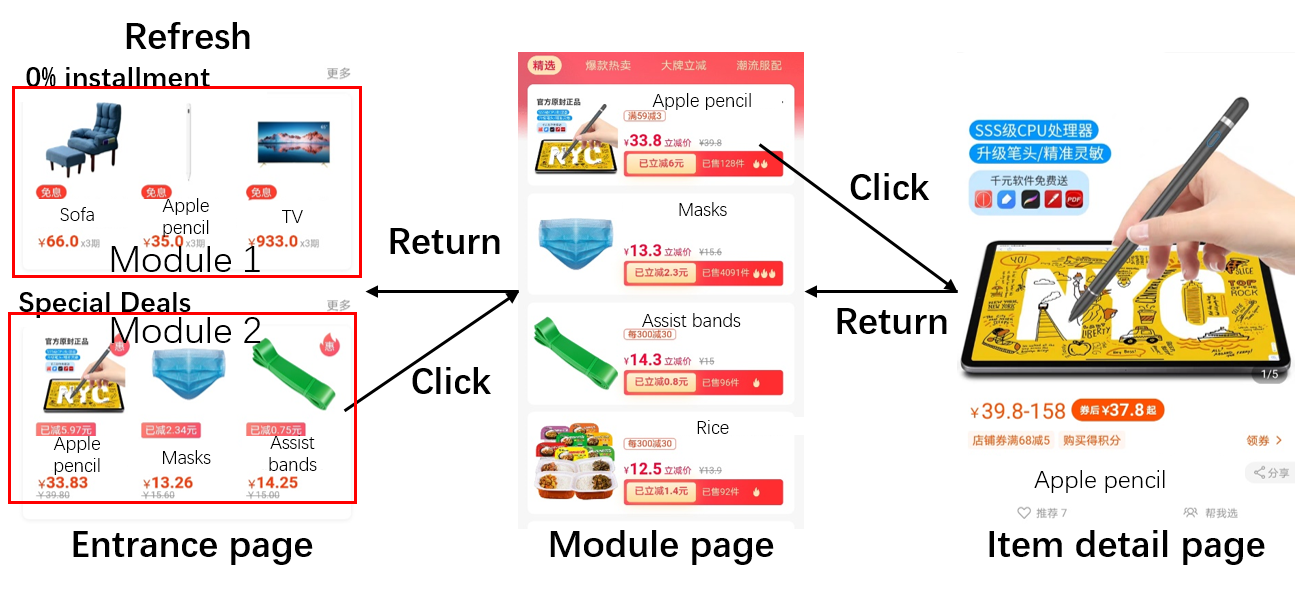}\\
  \caption{The flow of a multi-module recommendation system. Pages shown in this figure are the entrance page, the module page, and the item detail page. The entrance page contains two modules.}\label{flow}
\end{figure*}

We firstly introduce the details of the multi-module recommendation problem. Fig. \ref{flow} shows three stages in a recommendation session. 
First, when a user enters the recommendation scenario, he firstly browses the \textit{entrance page}, which contains more than one module. The ranking strategy of each module recommends items from its candidate set depending on users' information. A list of items is ranked and the top-3 will be shown on the entrance page. 
The user can 1) go to the \textit{module page} if he clicks any module, or 2) refresh the web page to access new items shown in modules, in which ranking strategies are called again to rank items.
Second, the \textit{module page} shows a list of recommended items for this module and the first three items are consistent with the items showing on the entrance page. 
The user can 1) slide the screen to browse more items, 2) go to the \textit{item detail page} by clicking an item, or 3) return to the entrance page. The agent will recommended more items if the whole list is browsed.
Third, the \textit{item detail page} demonstrates the details of an item. The user can 1) purchase the item, or 2) return to the module page.
The recommended items do not change when the user returns to the module page and he can continue to explore more preferred items by sliding the screen.
% Meanwhile, when a user returns to the entrance page, he can click another module or refresh the web page to access new items shown in modules, in which ranking strategies are called again to rank items.

Since different modules aim to collectively maximize the global performance, we can model our problem as a multi-agent extension of Markov Decision Processes (MDPs) \cite{littman1994markov}. 
Formally, the MDP for multiple agents is a tuple consisting of five elements $\langle N,S,A,R,P \rangle$:

% \textbf{Observation} $O$ is a set of observations  $[O_1,\dots,O_N]$ for each agent. In our problem, the observation for all the agents is different due to different candidate sets of items for different modules. The $i$-th agent's observation contains two parts: 1) The clicking history $h=[h_1,\dots,h_K]$ for a specific user, where $K$ is the length of the recorded history. 2) The information of the candidate items $S_n$ for $i$-th agent.
% $O_n=[h_1^i,\dots,h_T^i]$ consists of the clicking history $h$ and the information of optional items such as the distribution of prices and categories, where $T$ is the length of the history.
\textbf{Agent} $N$ is the number of agents. We treat modules as different agents rather than pages in existing works \cite{feng2018learning,DBLP:journals/corr/abs-1902-03987}.

\textbf{State} $S$ includes information that each agent has received about users. In our problem, $s$ is the information of users which contains: 1) static features such as age, gender, and address. 2) sequential features $[h_1,\dots,h_K]$ including features of $K$ items that a user purchased or clicked recently. 

\textbf{Action} $A=[A^1,\dots,A^N]$ is a set including the action sets of each agent. Specifically, $a = [a^1,\dots,a^N]$, where $a^i \in A^i$ is the action of the agent $i$. The action of each agent is defined as a weight vector that determines the rank of candidate items. Formally, the $j$-th element of the $i$-th agent's action $a^i=[a_1^i,\dots,a_j^i,\dots]$ is the weight of the $j$-th element of the item's feature. The weighted sum of the action and an item's feature determines the rank of the item, that is $score_{item} = a^T e_{item}$, where $e_{item}$ is the embedding of an item's features.

%the recommendation list containing $M$ items, which will be demonstrated in the $n$th module.

\textbf{Reward} $R = [R^1,\dots,R^N]$, where $R^i:S \times A \to \mathbb{R}$ is the reward function for agent $i$. After agents take action $a$ at the state $s$, the user would provide feedback like clicking an item or skipping the module, which can be converted to reward. The global reward $r$ will be obtained according to the reward function $r=R(s,a)$, where $r=[r^1,\dots, r^N]$ including rewards for $N$ agents.

\textbf{Transition probability} $P$ defines the probability $p(s_{t+1}|s_t,a_t)$ that the state transits from $s_t$ to $s_{t+1}$ given the action $a_t$ of all the agents in round $t$. In our setting, the transition probability is equivalent to user behavior probability, which is unknown and associated with $a_t$. The details are described in the experiment part.
% If the user clicks an item ranked by $a_t$, the sequential feature of the user will change from $[h_1,\dots,h_K]$ to $[h_2,\dots,h_{K+1}]$. 
% Thus, the state $s$ will also vary.

The \textbf{objective} of our problem is to maximize the discounted total reward of the platform $$\sum_{i=1}^N\sum_{t=0}^T \gamma^t r^i_t$$ rather than the independent reward of each agent $r^i_t$, where $T$ is the time horizon, and $\gamma^t$ is $t$-th power of the discounted parameter $\gamma$ to decide the weights of future rewards.

\begin{figure}[!t]
  \centering
  \includegraphics[width=.45\textwidth]{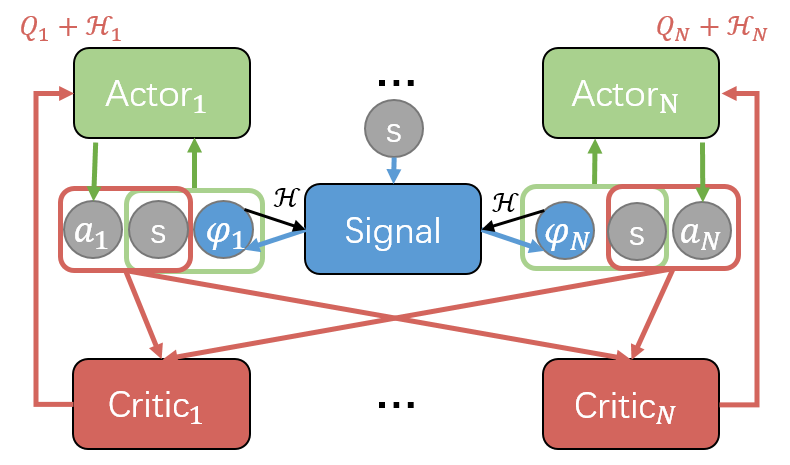}\\
  \caption{The architecture of our approach.  During the training, critics leverage other agents' actions to output the estimate of Q value. For the execution, each agent does not communicate with each other. }\label{signal}
\end{figure}

% \begin{figure}[!t]
%   \centering
%   \includegraphics[width=.40\textwidth]{embedding.png}\\
%   \caption{The architecture of state embedding.}\label{embedding}
% \end{figure}

% \begin{figure}[!t]
%   \centering
%   \includegraphics[width=.37\textwidth]{actor.png}\\
%   \caption{The architecture of actor and ranking.}\label{actor}
% \end{figure}

\section{Multi-Agent RL with a Soft Signal Network }
In this section, we propose a novel multi-agent reinforcement learning algorithm to address the multi-module recommendation problem. 
The main idea is to use a signal network to coordinate all the agents to maximize the global reward. Signals can be considered as the information of a general cooperative structure for all the agents. Then, agents act based on signals to cooperate.

Fig. \ref{signal} illustrates the structure of our algorithm, which is based on MADDPG \cite{lowe2017multi}.
Three components are involved in our structure. A shared signal network takes the state $s$ as input and sends signals $\phi$ to all the agents to maximize the overall performance. An actor maintained by each agent maps state and signal to action. The $i$-th actor-network only depends on the state and the signal for the $i$-th agent, without the knowledge of other agents. 
To estimate the expected future cumulative reward $Q^i(s,a)$ for given actions and states, each agent has a critic. 
In the centralized training, critics can evaluate the value of actions with information of all agents.
We describe the details of our model and training method in the following respectively.

% During decentralized execution, each agent keeps a copy of the signal network to generate its own signal.

\begin{figure*}[!t]
 \centering
 \subfigure[The architecture of state embedding. The features of users are divided into two types: static feature, and sequential feature. The attention mechanism is used to decide the weights of items recorded in the sequential feature. The embedding of state is not shared among different networks (including actors, critics and the signal network).]{
 \includegraphics[width=.58\textwidth]{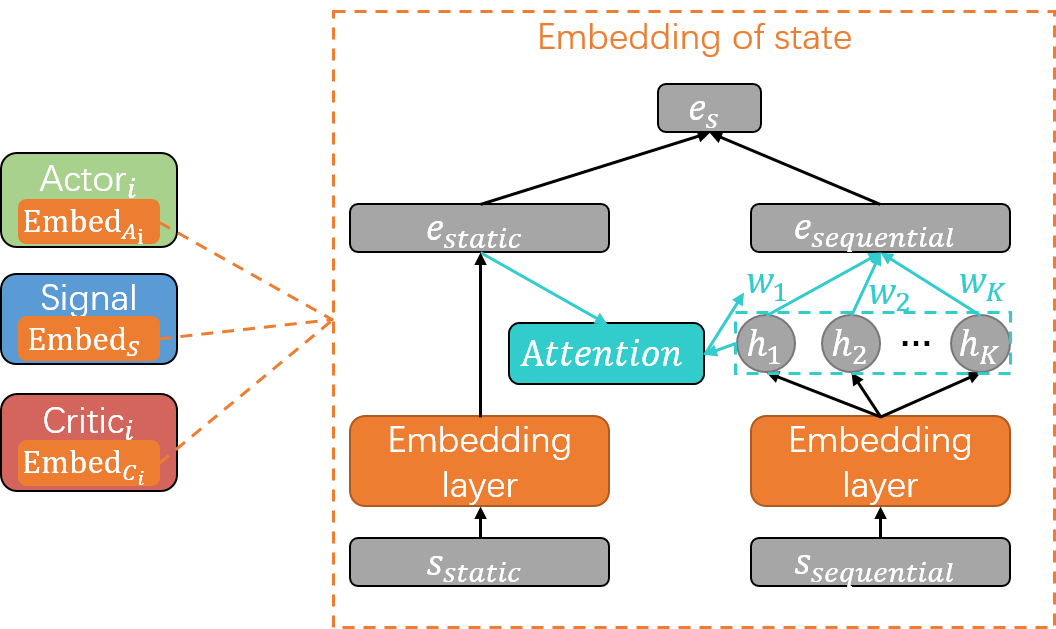}\label{embedding}
 } \hspace{0.2cm}
 \subfigure[The architecture of actor and ranking. The action is a vector whose dimension is the same with the item's feature. The weighted sum of these two vectors is considered as the score of an item.]{
 \includegraphics[width=.32\textwidth]{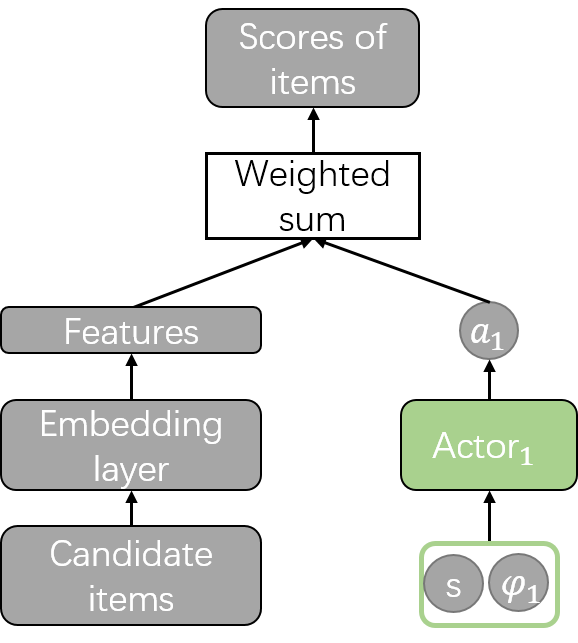}\label{actor}
 }
 \caption{State embedding and the structure of ranking.}
\end{figure*}

\subsection{Actor-Critic with a Signal Network}
\textbf{Embedding of state} We leverage the embedding layer and attention mechanism to extract useful information. The structure is shown in Fig. \ref{embedding}. As mentioned in Section \ref{form}, the state is the information of users that can be divided into two types, static and sequential features. For the static features like gender, each feature is processed by an independent embedding layer. While for the sequential features, $s_{sequential}$ includes different types' features of $K$ historical clicked items of a user such as item IDs and categories. Features belonging to one type share an embedding layer. For example, the item IDs of the 1st and the $K$-th items use the same layer. After embedding, sequential features are transformed to a set of vectors $h = [h_1,h_2,\dots,h_K]$, where $h_k$ is a vector containing the $k$-th item's features. We build an attention network to estimate the importance $w_k$ of $h_k$. The attention network takes the embedding of static information $e_{static}$ and $h$ as input. The outputs are mapped by the softmax function to obtain the regularized weights $w = [w_1,w_2,\dots,w_K]$, where $0\leq w_k \leq 1$ and $\sum_k w_k = 1$. The embedding of sequential features $e_{sequential}$ is generated by the weighted sum $\sum_k w_k h_k$. And it is concatenated with $e_{static}$ to get the embedding of the state $e_s$.
This embedding structure is included in actors, critics and the signal network to process the state. The parameters of embedding structures are not shared among different agents and components.

\textbf{Signal} The signal network $\Phi$ is shared by all agents during execution to maximize the overall reward. It maps state to a set of vectors $[\phi^1,\phi^2,\dots,\phi^N]$, where $\phi^i$ is the signal vector for the $i$-th agent. The state is processed by the embedding layer mentioned above and fully-connected layers output the signals depending on the embedding of state.
Differing from the communication mechanism that needs information sent by all agents, the signal network only depends on states. 
We adopt stochastic signal policies in which $\phi^i$ is sampled from a Gaussian distribution $\mathcal{N}(\mu_{\phi^i},diag(\sigma_{\phi^i}))$ where $[\mu_{\phi^i},\sigma_{\phi^i}]$ is the output of the signal network $$[\mu_{\phi^1},\sigma_{\phi^1},\mu_{\phi^2},\sigma_{\phi^2},\dots,\mu_{\phi^N},\sigma_{\phi^N}] = \Phi(s).$$
% This setting aims at encouraging exploration to avoid sub-optimal solutions.

\textbf{Actor} The structure of actors are illustrated in Fig. \ref{actor}. Each actor $\pi^i$ outputs an action given state $s$ and signal $\phi^i$. We concatenate the embedding of state and the signal as the input of a three-layer fully-connected network to generate action.
We define that the action of each module is a vector whose dimension is the same as the dimension of candidate items' features.
Following soft actor-critic \cite{haarnoja2018soft}, we adopt stochastic policy $[\mu_{a^i},\sigma_{a^i}] = \pi^i(s,\phi^i)$ and the action $a^i$ is sampled from $\mathcal{N}(\mu_{a^i},diag(\sigma_{a^i}))$.
To rank items, we leverage a linear model, in which the weighted sums of items' features and the action are treated as the scores of items. Candidate items are ranked and recommended according to their scores.

\textbf{Critic} Each agent maintains a critic network $Q^i(s,a)$ to estimate the expected cumulative reward of a state-action pair $(s,a)$. The embedding of the state is concatenated with actions of all the agents as the input of a fully-connected network whose output is $Q^i(s,a)$. In our problem, users' historical activities are collected and stored after users leave the multi-module scenario. During the training period, agents can access the actions of other agents to reduce the uncertainty of the environment. Since we use the soft actor-critic structure \cite{haarnoja2018soft}, the double-Q technique is adopted and a state value network $V(s)$ is maintained to stabilize training. The double-Q technique reduces the variance and over-fitting of the Q value by maintaining two Q networks and choosing the minimal Q value as the estimate of the $(s,a)$ pair in each time step. The value network $V(s)$ is used to approximate $\mathbb{E}_{s\sim\rho^{\pi},a\sim\pi}[\min Q_j(s,a)]$ for $j \in \{1,2\}$ and update Q networks.

\begin{algorithm}[t]
% \SetKwInOut{Initialize}{Initialize}
\SetKwInOut{Input}{Input}
\SetKwProg{Fn}{Function}{}{}
\caption{Multi-Agent Soft Signal-Actor (MASSA)}\label{masacss}
Initialize parameter vectors $(\theta,\eta,\tau,\xi,\delta)$, $\hat{\eta}=\eta$ \;
Initialize replay buffer $D$ \;
\For{$t = 0,1,\dots$}{
    Observe state $s_t$\;\label{exe0}
    For each agent $i$, generate signal $\phi^i = \Phi^i(s_t)$ and select action $a^i_t = \pi^i(s_t,\phi^i_t)$\;
    Execute action $a_t = [a^1_t,\dots,a^N_t]$ and observe reward $r_t$ and new state $s_{t+1}$\;
    Store $(s_t,a_t,r_t,s_{t+1})$ in the replay buffer $D$\;\label{exe1}
    Sample a batch of samples from $D$\;
    \For{each agent $i$}{
        Calculate $\nabla_\eta J^i_V(\eta)$ and update $\eta$\;
        Calculate $\nabla_{\theta_j} J^i_Q({\theta_j})$ and update $\theta_j$ for $j \in \{1,2\}$\;
        Calculate $\nabla_\tau J^i_\pi(\tau)$ and update $\tau$\;
    }
    Calculate $\nabla_\xi J^i_\Phi(\xi)$ and update $\xi$\;
    Update the parameter of the target state value network $\hat{\eta}_{t+1} = (1-\delta) \hat{\eta}_{t}+\delta\eta_{t}$\;
} 
\end{algorithm}
\subsection{Policy Update with the Soft Signal Network}
As discussed above, we use neural networks to approximate Q value, V value, represent policies and generate signals. For the $i$-th agent, we consider a parameterized state value function $V_\eta^i(s_t)$, two Q-functions $Q_{\theta_j}^i(s_t,a_t), \ j \in \{1,2\}$, a stochastic policy $\pi^i_\tau(s_t,\phi^i_t)$ and a shared signal network $\Phi_\xi(s_t)$. The parameters of these networks are $\eta$, $\theta$, $\tau$, and $\xi$. The update rules will be introduced in this section.

We adopt Soft Actor-Critic (SAC) \cite{haarnoja2018soft} for each agent.
Differing from standard RL that maximizes the expected sum of rewards $$\mathbb{E}_{s\sim\rho^{\pi},a\sim\pi}\left[\sum_{t=0}^T \gamma^t r_t\right],$$ the objective of SAC augments the objective with the expected entropy of the policy $\mathbb{E}_{s\sim\rho^{\pi},a\sim\pi}\left[\sum_{t=0}^T \gamma^t (r_t+\mathcal{H}(\pi^i(\cdot|s_t,\phi^i_t)))\right]$, where $\rho^{\pi}$ is the state distribution induced by $\pi$, $\gamma^t$ is the $t$-th power of $\gamma$ and $\mathcal{H}(\pi^i(\cdot|s_t,\phi^i_t) = \mathbb{E}_{a_t^{i}\sim\pi^i}\left[\pi^i_\tau(a^i_t|s_t,\phi^i_t)\right]$. The entropy term aims at encouraging exploration, while giving up on clearly unpromising avenues. 
We have 
\begin{equation}\label{q}
    Q^i(s_t,a_t) = \mathbb{E}_{s\sim\rho^{\pi},a^i\sim\pi^i}\left[r(s_t,a_t) + \gamma V^i(s_{t+1})\right],
\end{equation}
where $$V^i(s_{t}) = \mathbb{E}_{a^i\sim\pi^i}\left[ Q^i(s_t,a_t)-\log \pi^i(a_t^i|s_t,\phi^i_t)\right].$$ Then, we update the parameters of $Q$ and $V$ according to \cite{haarnoja2018soft}.

\textbf{Critic.} The centralized critic is optimized according to the Bellmen function of soft actor-critic. For the value function $V^i_\eta(s_t)$, we have
\begin{equation}
    J^i_V(\eta) = \mathbb{E}_{s_t\sim D}\left[\frac{1}{2}\left(V_\eta^i(s_t)-\mathbb{E}_{a_t\sim \pi_\tau}\left[Q_\theta^i(s_t,a_t)-\log \pi^i_\tau(a^i_t|s_t,\phi^i_t) \right]\right)^2\right],
\end{equation}
where $D$ is the distribution of samples, or a replay buffer. The gradient of $i$-th V value network can be estimated by an unbiased estimator:
\begin{equation}
    \hat{\nabla}_\eta J^i_V(\eta) = \nabla_\eta V_\eta^i(s_t) \left(V_\eta^i(s_t)-Q_{\theta}^i(s_t,a_t)+ \log \pi^i_\tau(a^i_t|s_t,\phi^i_t) \right),
\end{equation}
where actions and signals are sampled from current networks and $Q_{\theta}^i = \min_{j \in \{1,2\}} Q_{\theta_j}^i(s_t,a_t)$. The Q value network is trained to minimize the Bellman residual
\begin{equation}
    J^i_Q(\theta_j) = \mathbb{E}_{(s_t,a_t)\sim D}\left[\frac{1}{2}\left(Q_{\theta_j}^i(s_t,a_t)-\hat{Q}^i(s_t,a_t)\right)^2\right],
\end{equation}
with 
\begin{equation}
    \hat{Q}^i(s_t,a_t) = r_t^i(s_t,a_t) + \gamma \mathbb{E}_{s_{t+1}\sim P}\left[V^i_{\hat{\eta}}(s_{t+1})\right],
\end{equation}
where $V^i_{\hat{\eta}}(s_{t+1})$ is a target network of $V$, where $\hat{\eta}$ is an exponentially average of $\eta$. More specifically, the update rule for $\hat{\eta}$ is $\hat{\eta}_{t+1} = (1-\delta)\hat{\eta}_{t} + \delta \eta_{t}$.
We approximate the gradient for $\theta_j$ with
\begin{equation}
    \hat{\nabla}_{\theta_j} J^i_Q(\theta_j) = \nabla_{\theta_j} Q_{\theta_j}^i(s_t,a_t) \left(Q_{\theta_j}^i(s_t,a_t)-\hat{Q}^i(s_t,a_t)\right).
\end{equation}
\textbf{Actor.} For each actor, the objective is to maximize the Q value with the entropy term, since Q value introduced in Eq. (\ref{q}) does not include $\mathcal{H}(\pi^i(\cdot|s_t,\phi^i_t)$:
\begin{equation}
    J^i_\pi(\tau) = -\mathbb{E}_{s_t\sim D, a^i\sim \pi^i}\left[Q_\theta^i(s_t,a_t^{-i},a_t^i)-\log \pi^i_\tau(a^i_t|s_t,\phi^i_t)\right],
\end{equation}
where $a_t^{-i}$ is a vector including actions of all the agents except the $i$-th agent and $a_t^i$ is generated by the current policy $\pi^i_\tau$. We use reparameterization trick \cite{kingma2013auto} $$a^i_t = f_\tau(\epsilon_t;s_t,\phi^i_t) = f^\mu_\tau(s_t,\phi^i_t) + \epsilon f^\sigma_\tau(s_t,\phi^i_t),$$ where $\epsilon \sim \mathcal{N}(0,I)$ and $I$ is identity matrix. $f_\tau$ is a neural network whose output is $[f^\mu_\tau,f^\sigma_\tau]$ and $\tau$ is the parameter of $f_\tau$.
The stochastic gradient is 
\begin{equation}
\begin{split}
    \hat{\nabla}_{\tau}J^i_\pi(\tau) =& \nabla_{\tau} \log \pi^i_\tau(a^i_t|s_t,\phi^i_t) + \left(-\nabla_{a_t^i}Q_\theta^i(s_t,a_t^{-i},a_t^i)+ \right.\\
    &\left.\nabla_{a_t^i} \log \pi^i_\tau(a^i_t|s_t,\phi^i_t)\right) \nabla_{\tau}f_\tau(\epsilon_t;s_t,\phi^i_t).
\end{split}
\end{equation}
These updates are extended from soft actor-critic algorithm \cite{haarnoja2018soft}. 

\textbf{The entropy-regularized signal network.}
Now we introduce the update of the signal network. Since the signal network aims at maximizing the overall reward, the objective function is 
\begin{align}
    J_\phi(\xi) &= \frac{1}{N}\sum_i -\mathbb{E}_{s_t,a_t^{-i}\sim D}\left[Q_\theta^i(s_t,a_t^{-i},a_t^i)\right].
    % \sum_i J^i_\pi(\xi) \\
    % & = \sum_i \mathbb{E}_{s_t,a_t^{-i}\sim D}\left[-Q_\theta^i(s_t,a_t^{-i},a_t^i)+\log \pi^i_\tau(a^i_t|s_t,\phi^i_t)\right]
\end{align}
Inspired by the soft actor-critic, we augment an expected entropy of the signal network (soft signal network) and obtain a new objective. Intuitively, this term can encourage signal network to coordinate agents' exploration and find the optimal solution to maximize the global reward. Since the signal network outputs a signal $\phi^i$ for each agent $i$, we use the notation $\Phi^i$ to represent the part of the signal network for the $i$-th agent. Since $$\mathcal{H}(\Phi^i(\cdot|s_t)) = \mathbb{E}_{\phi^i \sim \Phi^i}\log \Phi^i(\phi^i|s_t),$$ we have

\begin{equation}
\begin{split}
    J_\phi(\xi) = & \frac{1}{N} \sum_i \left[\mathbb{E}_{s_t,a_t^{-i}\sim D, \phi^i \sim \Phi^i}\left[-Q_\theta^i(s_t,a_t^{-i},\pi^i(s_t,\phi_t^i))+ \right.\right. \\ 
    &\left.\left. \alpha\log \Phi^i(\phi_t^i|s_t)\right] \right].
\end{split}
\end{equation}

% \begin{align}
%     J_\phi(\xi) &= \frac{1}{N}\sum_i \left[-\mathbb{E}_{s_t,a_t^{-i}\sim D}\left[Q_\theta^i(s_t,a_t^{-i},\pi^i(s_t,\phi^i))\right] - \alpha \mathcal{H}(\Phi^i(\cdot|s_t))\right] \\
%     &= \frac{1}{N}\sum_i \left[-\mathbb{E}_{s_t,a_t^{-i}\sim D}\left[Q_\theta^i(s_t,a_t^{-i},\pi^i(s_t,\phi^i))\right] + \alpha \mathbb{E}_{\phi^i \sim \Phi}\log \Phi^i(\phi^i|s_t)\right]\\
%     &=\frac{1}{N} \sum_i \left[\mathbb{E}_{s_t,a_t^{-i}\sim D, \phi^i \sim \Phi}\left[-Q_\theta^i(s_t,a_t^{-i},\pi^i(s_t,\phi_t^i))+ \alpha\log \Phi^i(\phi_t^i|s_t)\right] \right]
% \end{align}
According to \cite{jankowiak2018pathwise}, we derive the stochastic gradient using reparameterization trick again $\phi^i_t = g^i_\xi(\epsilon;s_t) = g^{\mu}_\xi(s_t) + \epsilon g^\sigma_\xi(s_t)$, where $g$ is a neural network and $\epsilon \sim \mathcal{N}(0,I)$:
% \alpha \nabla_{\xi} \log \Phi^i(\phi_t^i|s_t) +
\begin{equation}
\begin{split}
    \hat{\nabla}_{\xi} J_\Phi(\xi) &= \frac{1}{N}\sum_i \left[ \alpha \nabla_{\xi} \log \Phi^i_\xi(\phi_t^i|s_t) + \left( \alpha \nabla_{\phi^i_t} \log \Phi^i(\phi_t^i|s_t) - \right. \right. \\ 
    &\left.\left. \nabla_{a_t^i} Q_\theta^i(s_t,a_t^{-i},a_t^i)\nabla_{\phi^i_t} \pi^i_\tau(a_t^i|s_t,\phi^i_t) \right) \nabla_{\xi} g^i_\xi(\epsilon_t;s_t) \right].
\end{split}
\end{equation}

The whole algorithm is shown in Algorithm \ref{masacss}. The algorithm can be divided into two stages, execution and training. In the execution part (Lines \ref{exe0}-\ref{exe1}), policies of different agents are executed in the environment to collect data that is stored in the replay buffer. In the training part, all the parameters are updated according to their gradients derived in this section. In the end, the parameter of the target network is updated.

% For our problem, the ideal solution is that all the modules' rewards increase by cooperation.
% Inspired by the correlated equilibrium, we introduce a signal network $\mu$ shown in Fig. \ref{signal} based on the idea of hierarchical policy to find the optimal policy. The input of the signal network $\mu$ is the observation $o$ and parameters $\theta$ of the policy network of all the agents. The signal network can be seen as a centralized controller that knows all the agents' information clearly. Then signal $z_i$ is generated for each agent $i$. We hope that the signal $z_i$ can help individuals to choose an action that maximizes the global reward $r_g$ which is approximated by a global critic network $Q_\phi(s,z)$. Thus, we use the $Q_\phi(s,z)$ to train the signal network and set the loss as $-Q_\phi(s,z)$. For each agent's policy network $\pi^i_\theta(s,z_i)$, the objective is to maximized the local return $Q_i(s,z,a) = \sum_{t=0}^T \gamma^t r^t_i$.

\begin{algorithm}[t]
% \SetKwInOut{Initialize}{Initialize}
\SetKwInOut{Input}{Input}
\SetKwProg{Fn}{Function}{}{}
\caption{Offline testing procedure.}\label{offline_test}
Load parameters of actors, signal network and item embedding layer\;
\For{$t = 0,1,\dots$}{
    Read a record from testing dataset\;
    Observe state $s_t$\;
    Observe candidate set of items for two modules $L^i, \ i \in {1,2}$\;
    For each agent, rank these items and output a list\;
    Observe rewards $r_t$ of recommended lists from the record\;\label{reward}
    Generate next state $s_{t+1}$ (for training only)\;
}
\end{algorithm}

\begin{figure*}[!t]
\centering
  \includegraphics[width=.6\textwidth]{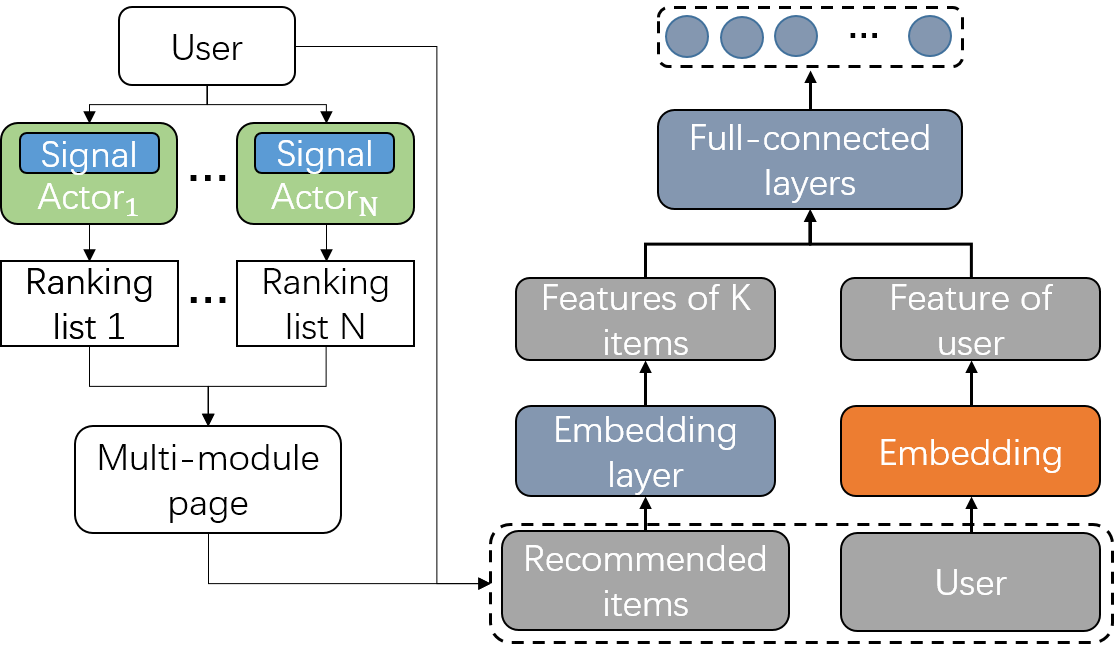}\\
  \caption{The structure of our simulator. The inputs are all the recommended items on one web page and the information of a user. The output is a vector including the probabilities that these items are clicked.}\label{simulator}
\end{figure*}
\begin{table*}[!t]
    \centering
    \caption{Results of offline testing.}
    \scalebox{1.1}{
    \begin{tabular}{c|c|c|c|c|c|c}
    \hline
    \hline
    \multirow{2}{*}{\diagbox[]{Method}{Metric}} & \multicolumn{3}{c|}{Precision}&\multicolumn{3}{c}{nDCG}\\
    \cline{2-7}
    & Module 1 & Module 2 & Overall & Module 1 & Module 2 & Overall\\
    \hline
    L2R& 0.193 & 0.047 & 0.24 & 0.196 & 0.042 & 0.238\\	
    % \hline
    DDPG& 0.211 & 0.046 & 0.257 & 0.214 & 0.042 & 0.256\\
    \hline
    MADDPG& 0.227 & 0.047 & 0.274 & 0.231 & 0.044 & 0.275\\
    % \hline
    COMA& 0.165 & 0.039 & 0.204 &0.175  &0.037  & 0.212\\
    % \hline
    QMIX& 0.396 &0.056 & 0.452 & 0.368 &0.055  & 0.423\\
    % \hline
    COM& 0.216 &0.042 & 0.258&0.217 &0.041 &0.258\\ 						
    \hline
    MASAC& 0.367 & 0.055 & 0.422 & 0.337 & 0.051 & 0.389\\
    % \hline
    COMA+SAC& 0.206 & 0.048 & 0.254 &0.205  &0.045  & 0.25\\ 
    % \hline
    QMIX+SAC& 0.305 &0.048 & 0.353 & 0.294 &0.042  & 0.336\\
    % \hline
    COM+SAC& 0.292 &0.047 & 0.341&0.29 &0.045 &0.335\\
    % \hline
    MAAC& 0.301 & 0.046 & 0.347 & 0.289 & 0.043 & 0.332\\			
    \hline
    MASSA w/o att (ours) &0.433&0.052&0.485&0.397&0.050&0.447\\
    MASSA w/o en (ours)& 0.44  &0.055 & 0.495 & 0.398& 0.05  & 0.448\\
    % \hline
    MASSA (ours)& \textbf{0.555} &\textbf{0.06}  & \textbf{0.615} & \textbf{0.459} &\textbf{0.057}  & \textbf{0.516}\\						
    \hline
    \hline
    \end{tabular}}
    \label{offline_result}
\end{table*}

\section{Experiment}
We conduct extensive experiments to evaluate the performance of our algorithm based on Taobao. We first describe the details of the experimental setting. Then, some baselines are introduced. Finally, the performance of baselines and our algorithm are illustrated.
\subsection{Dataset}
Our dataset is collected from Taobao. The recommendation scenario contains two modules. For the training data, 14-day data is collected in March 2020 and about 1.5 million records (583076 items) are included in the dataset. Another 3-day data (about 200 thousand records) is used as the test dataset in offline testing. 
Each record includes a user's information, 10 recommended items for each module, the user's clicks, and the user's information after clicking. 
As we mentioned in the formulation section, users' information contains sequential and static features. Sequential features contain 50 items that the user clicked. The item ID, seller ID, and category ID of these historical clicked items are stored. If the number of historical clicked items of a user is less than 50, these features are set to 0 by default.
For recommended items, features include price, sale, category and other information of items. After embedding, each item is represented by a 118-dimensional vector. 

\subsection{Experiment Setting}
In the experiment, we use both offline and online (simulator) testing to illustrate the performance of our algorithm. 
In the offline training and evaluation, algorithms re-rank browsed items in each record and the clicked items should be ranked at the top of the list. The candidate set is limited to the recommended items stored in each record rather than all items collected from the dataset since we do not know the real reward of items that the user does not browse. The rewards of the recommended lists of our algorithm are directly obtained from historical data and used to evaluate the performance of algorithms. During training, since we need $s_{t+1}$ to update parameters, the users' information $s_t$ is updated by following rules. The static part of $s$ is fixed and not changed no matter what the user clicks. If an item is clicked, the item is added into the sequential feature and the $50$-th historical clicked item is removed from the sequential feature. If $M$ item is clicked, we do $M$ updates of the sequential feature. We assume that items in the first module are clicked firstly and the items with high ranks are clicked before those with low rank within a module. Then, the new $s_{t+1}$ is generated and stored to train our algorithm. The offline testing algorithm in detail is presented in Algorithm \ref{offline_test}.

For the online training and testing, due to the huge cost and risk caused by deploying different algorithms to the real-world scenario, we train a simulator to implement the online testing following \cite{zhao2018deep}. The structure of the simulator is shown in Fig. \ref{simulator}. In order to consider the information on the whole page, the input of the simulator is all the recommended items of two modules (6 items). We obtain embedding of items by a shared embedding layer. Meanwhile, the user's information is processed by the embedding structure shown in Fig. \ref{embedding}. The features of items and a user are concatenated as the input of a four-layer fully-connected network. Then, the CTRs (Click Through Rate) of these items are predicted. The bias of position is considered by this design since the sequence of items in the input actually indicates the information of positions. We test the trained simulator in the test dataset (not used to train the simulator). The overall accuracy is over 90\%, which suggests that the simulator can accurately simulate the real online environment.

For training and testing our algorithm, we collect 2000 items with the largest CTR for each module to expand the candidate set. In our training and testing dataset, about 90\% of clicks are contributed by these items. In each round, actors select a list of items and the simulator outputs rewards for these items. The training and testing procedure is similar to Algorithm \ref{offline_test} except the Line \ref{reward}, where the rewards come from the simulator rather than historical data.

To evaluate the performance of various algorithms, we use clicks as rewards and introduce two metrics Precision \cite{manning2010introduction} and nDCG \cite{wang2013theoretical}. The formulations are shown as follows.
\begin{itemize}
    \item Precision: $$Precision = \frac{\text{\#clicks in top-K items}}{K}.$$
    \item nDCG: $$nDCG = \sum_{k=1}^K \frac{r_k}{\log(1+k)},$$ where $r_k = 1$ if the $k$-th item is clicked, otherwise, $r_k = 0$.
\end{itemize}
For each module, the performance of a ranking policy is evaluated by these two metrics. The overall performance is the sum of each module's performance.

For components of our algorithm, we leverage a 4-layer neural network with the additional embedding structure introduced in Fig. \ref{embedding}. The activation function is $relu$ for all fully-connected layers except output layers. The size of the replay buffer is $1e6$. The dimension of the items' embedding is $118$. The length of each signal vector is $64$. The discount factor is $\gamma=0.99$. The learning rate for actor, critic, and signal networks is $0.01$ and the weight for updating the target network is $\delta=0.01$. The weight of entropy terms is $\alpha=0.01$. We select these parameters via cross-validation and do parameter-tuning for baselines for a fair comparison.
\begin{figure*}[!t]
 \centering
 \subfigure[Precision of Module 1]{
 \includegraphics[width=.3\textwidth]{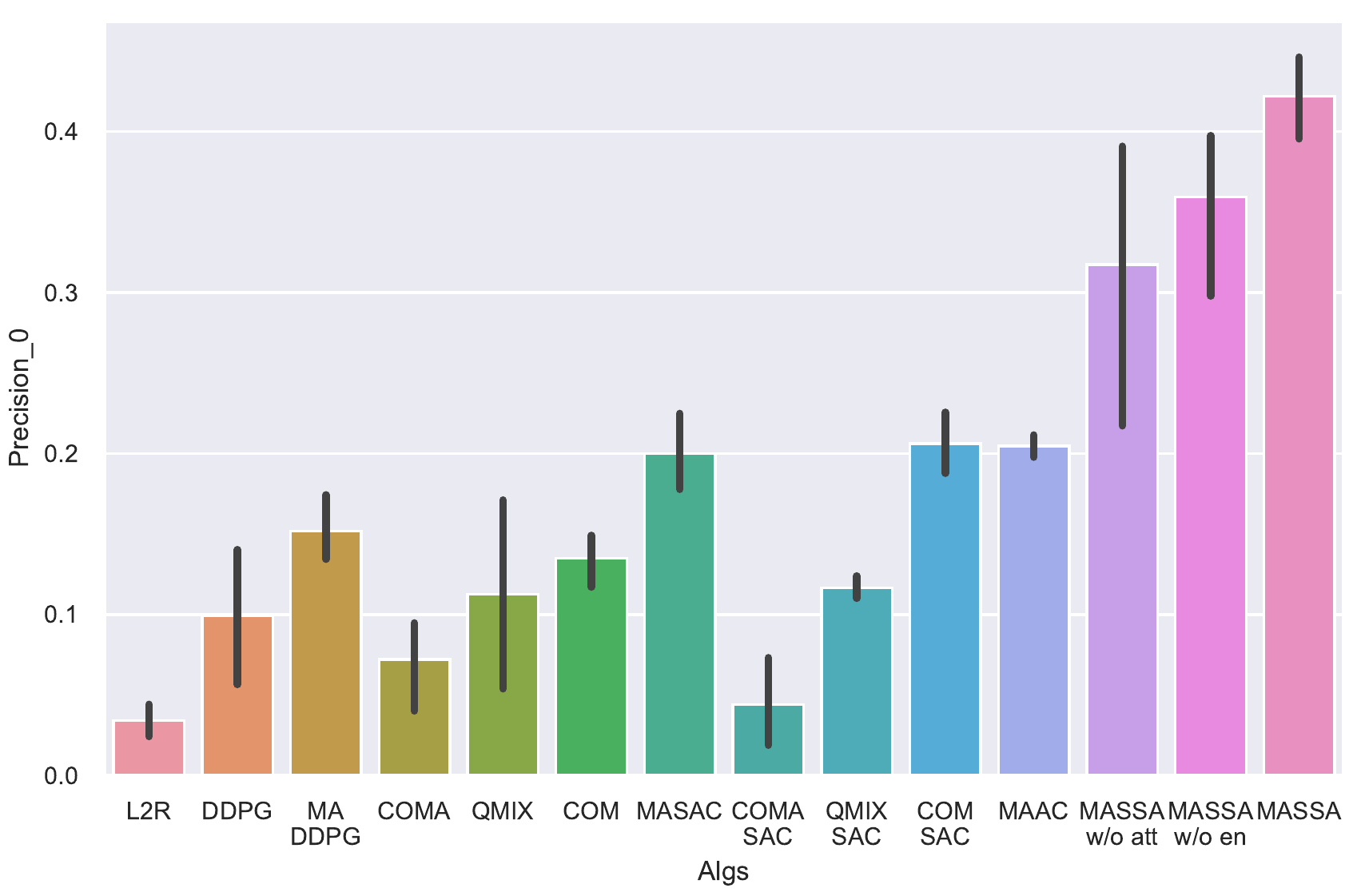}
 }
 \subfigure[Precision of Module 2]{
 \includegraphics[width=.3\textwidth]{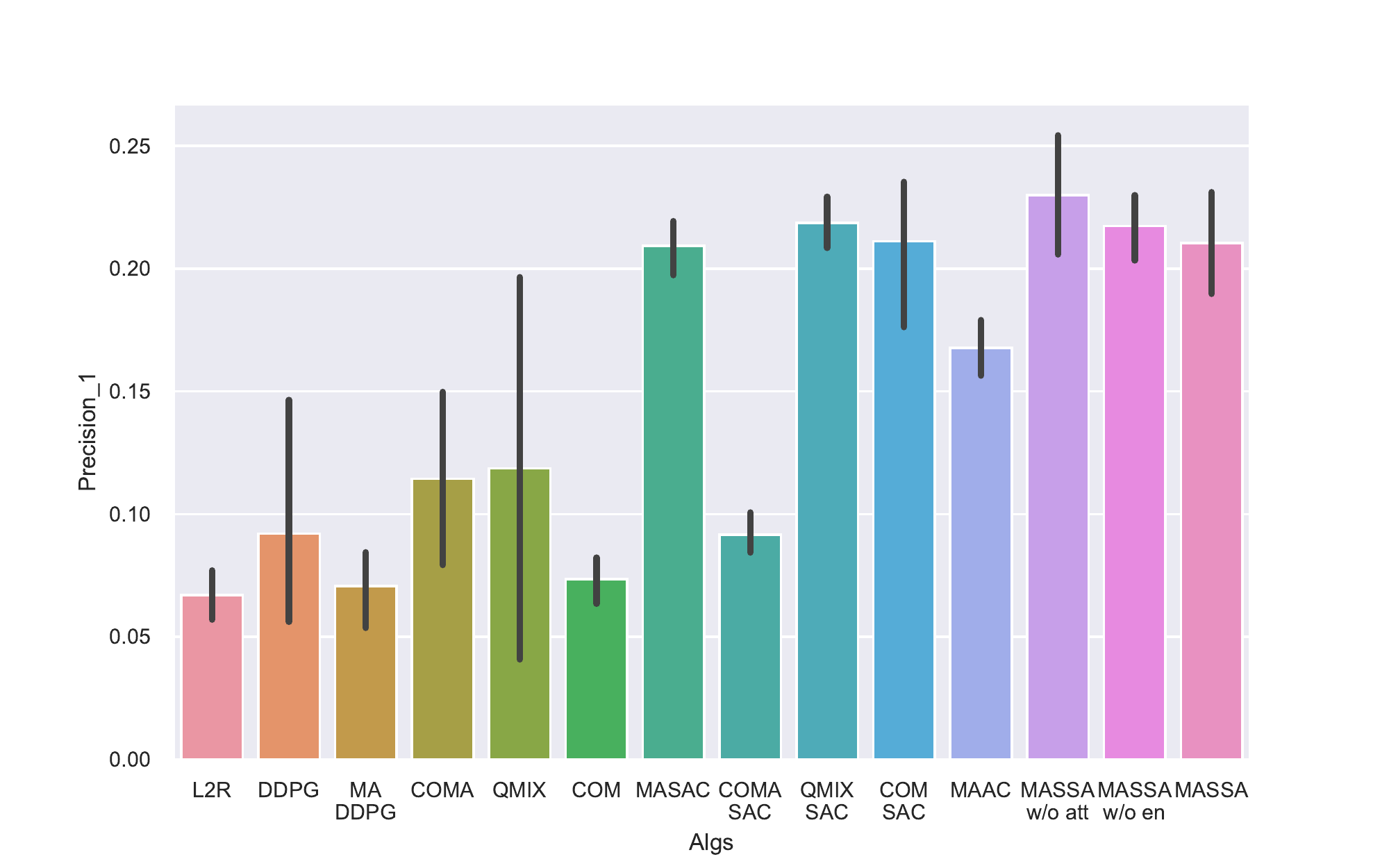}
 }
 \subfigure[Overall Precision]{
 \includegraphics[width=.3\textwidth]{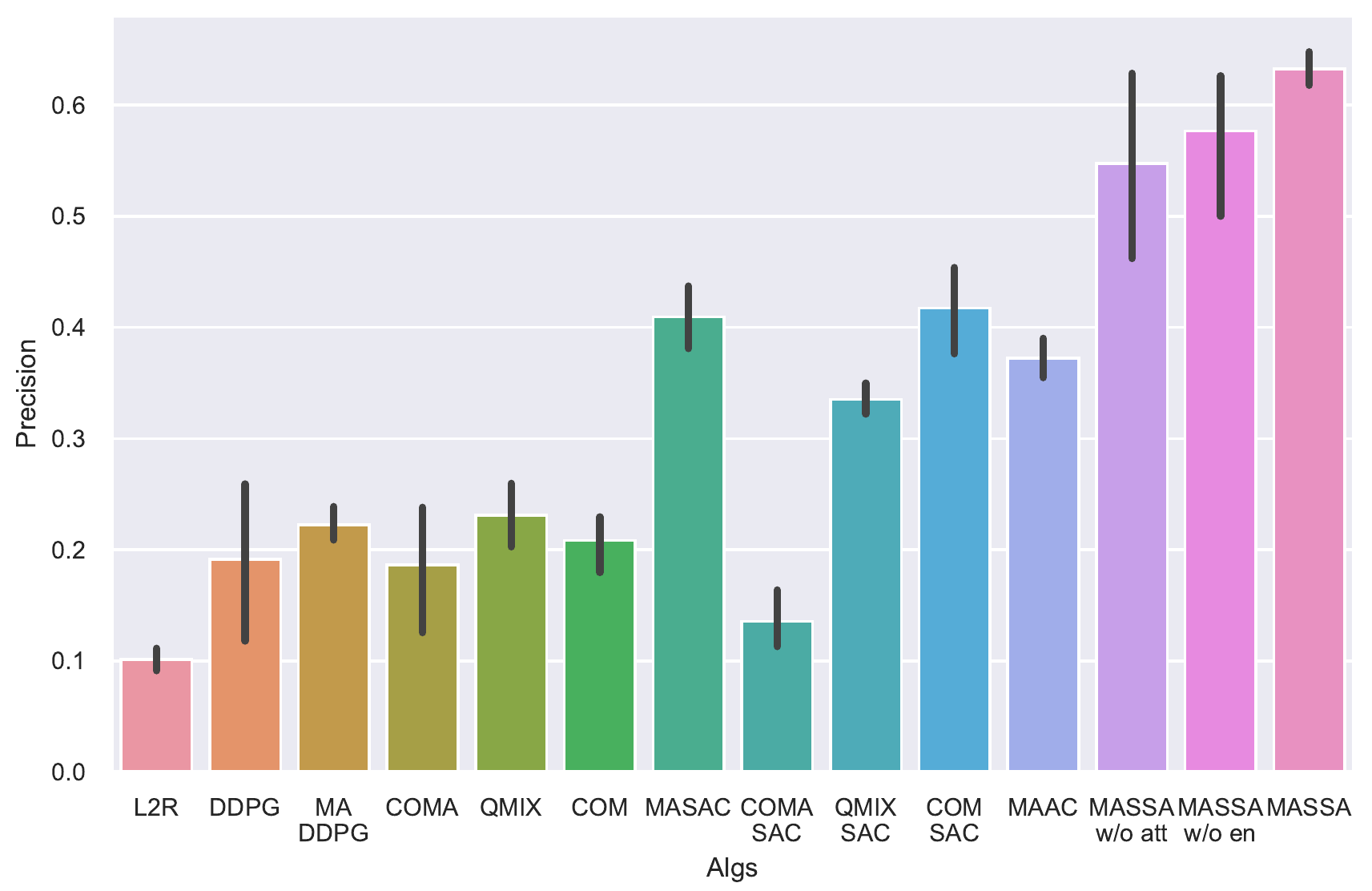}
 } \\
 \subfigure[nDCG of Module 1]{
 \includegraphics[width=.3\textwidth]{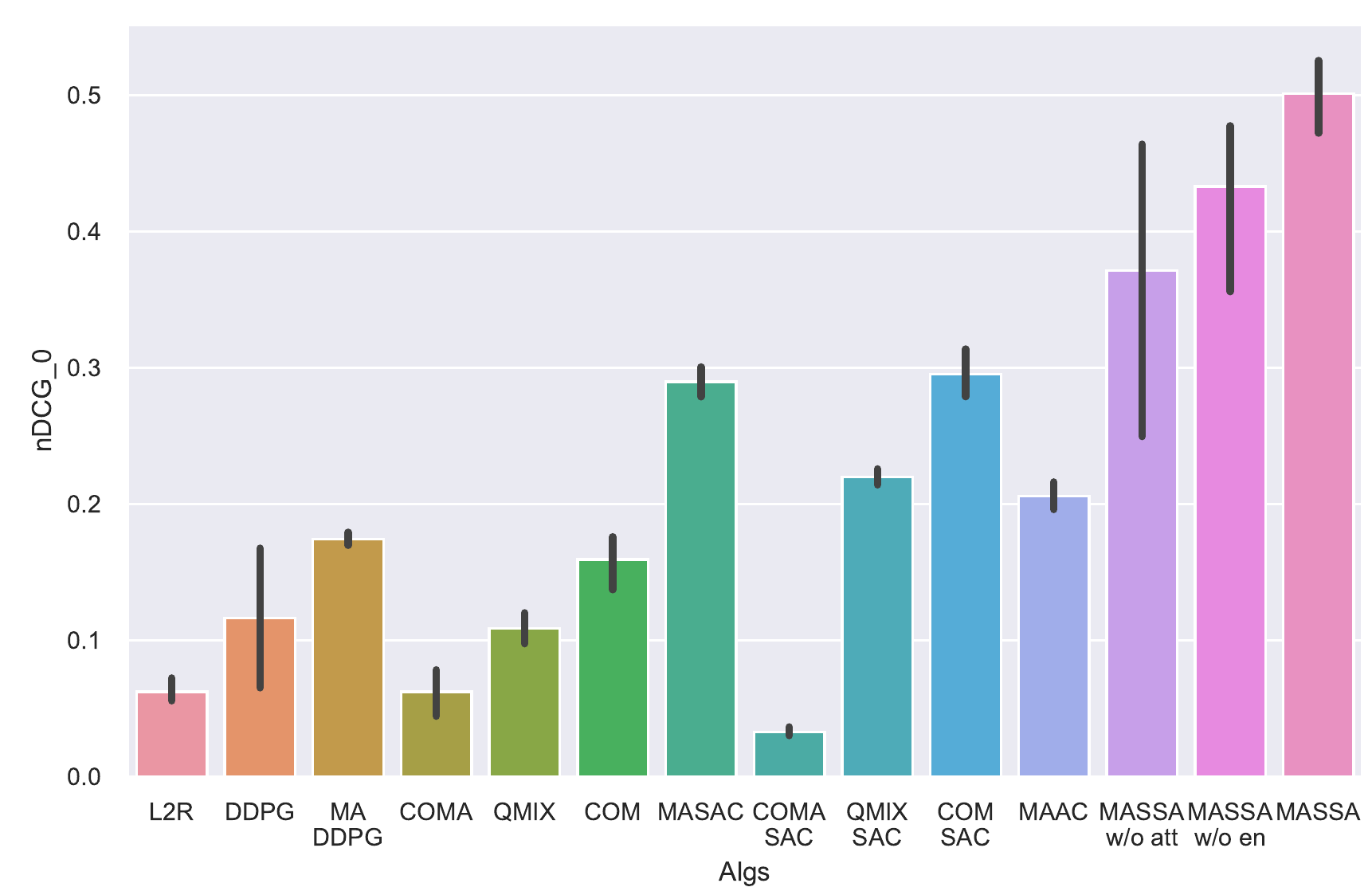}
 }
 \subfigure[nDCG of Module 2]{
 \includegraphics[width=.3\textwidth]{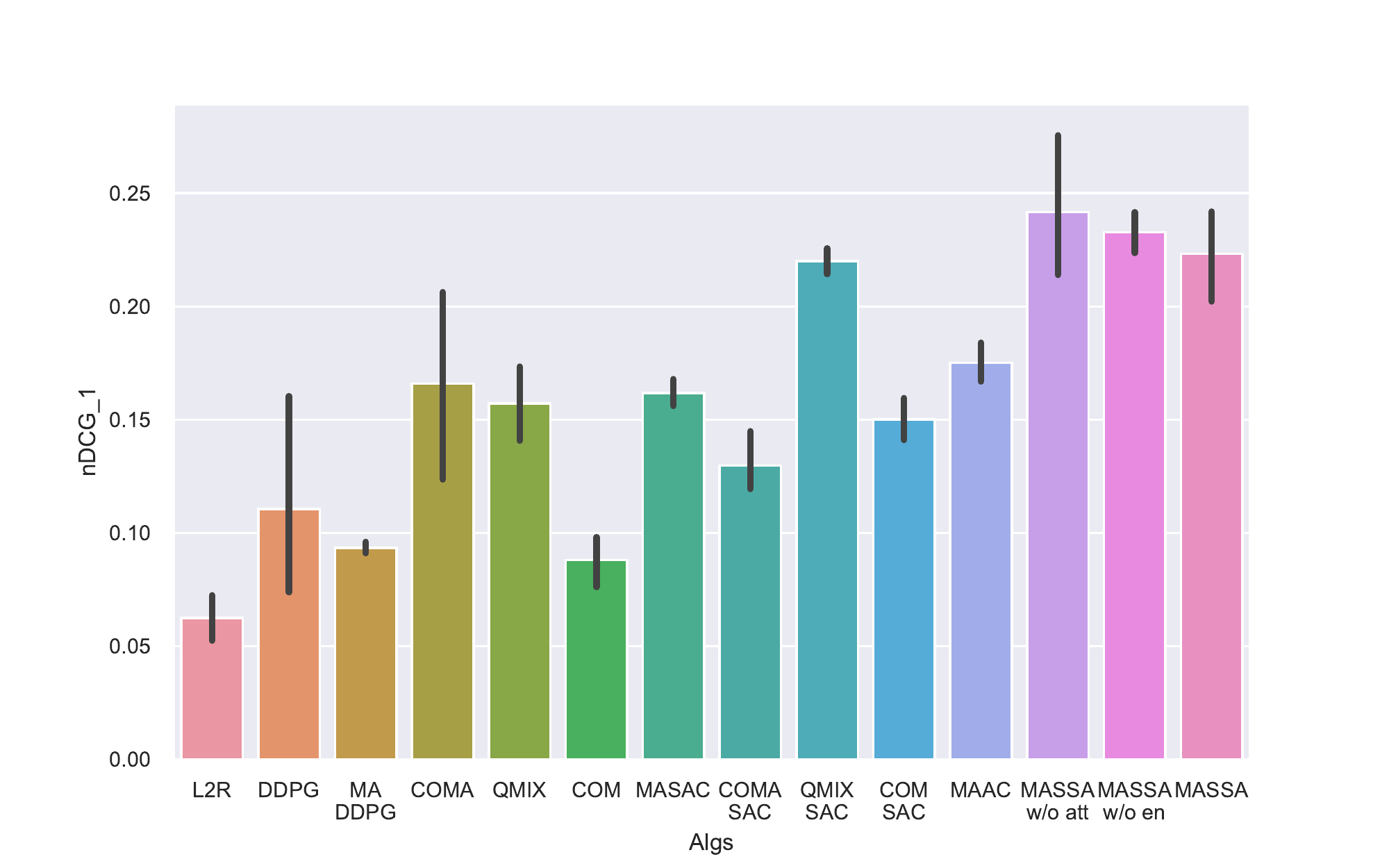}
 }
 \subfigure[Overall nDCG]{
 \includegraphics[width=.3\textwidth]{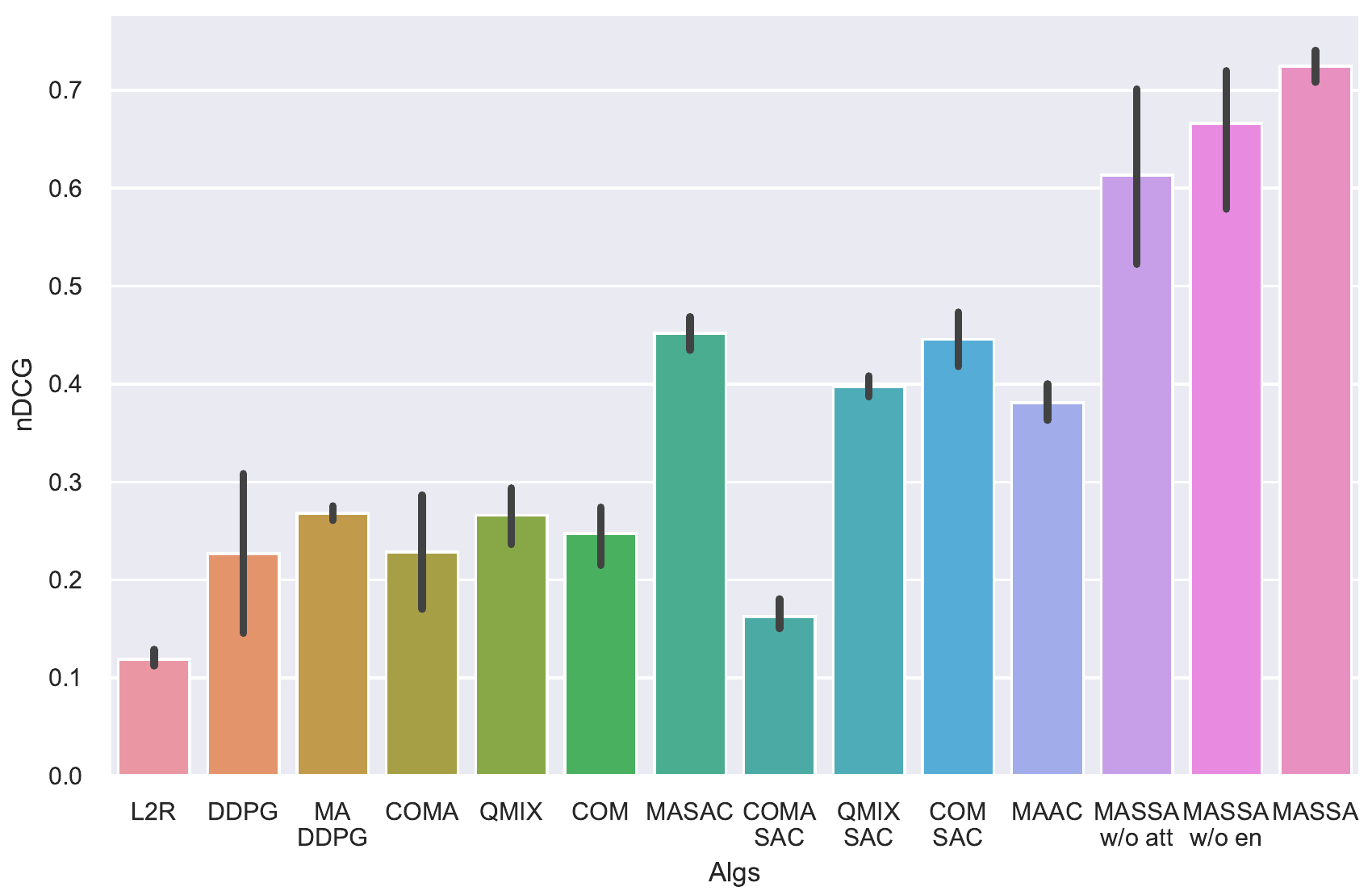}
 }
 \caption{The results of the online experiment.} \label{result_online}
\end{figure*}
\subsection{Baselines}
Our algorithm is compared with the following baselines:
\begin{itemize}
    \item \textbf{L2R} \cite{liu2009learning}: This algorithm trains a point-wise learning-to-rank network by supervised learning. The network is the same as the simulator except for the input and the output. The input changes to users' information and one item. The network predicts the CTR of this item. We deploy an L2R algorithm for each module, which is trained to reduce the sigmoid cross-entropy loss for each module.    
    \item \textbf{DDPG} \cite{lillicrap2015continuous}: Deep Deterministic Policy Gradient method is a single-agent RL algorithm that consists of an actor and a critic. The structure of actors is the same as MADDPG and L2R.
    \item \textbf{MADDPG} \cite{lowe2017multi}: Multi-Agent Deep Deterministic Policy Gradient method is the multi-agent version of DDPG. Each agent maintains an actor and a critic. During training, critics can access other agents' actions and observations. While in the execution, actors select actions only depending on their own observation. 
    \item \textbf{COMA} \cite{foerster2017counterfactual}: Counterfactual multi-agent policy gradients method is a cooperative multi-agent algorithm that leverages counterfactual rewards to train agents. The main idea is to change the action of an agent to a baseline action and use the gap of Q values of these two actions as the reward. Differing from MADDPG, all the agents share a critic to estimate the global reward.
    \item \textbf{QMIX} \cite{rashid2018qmix}: QMIX assumes that the global maximum reward is a weighted sum of local maximum rewards of agents and proposes a mixing network to explicitly decompose the global reward. The decomposed local rewards are treated as the contribution of each agent and used to train actors.
    \item \textbf{COM}: COM is a simple extension of the methods \cite{feng2018learning} by letting actors choose actions simultaneously. Actors send messages to others during execution. Although this algorithm violates the restriction that different modules cannot communicate, the comparison aims to illustrate the performance in environments that allow communication.
    \item \textbf{MASAC}: This algorithm is an extension of MADDPG by applying soft actor-critic \cite{haarnoja2018soft}, where an entropy term is augmented in the reward to encourage exploration. Different from our method, this algorithm does not have a signal network.
    \item \textbf{MAAC} \cite{iqbal2018actor}: Multi Actor-Attention-Critic algorithm maintains the structure of MASAC. The attention mechanism is adopted to handle messages sent by critics and extract useful information to each critic.
    \item \textbf{MASSA w/o en}: This method is proposed for the ablation study, in which the entropy terms of signals are removed from the loss function of the signal network ($\alpha=0$). By comparing this method with ours, the importance of the entropy-regularized version of the loss function is indicated.
    \item \textbf{MASSA w/o att}: In this method, the attention mechanism is replaced by simple concatenation $e_s = [e_{static},h]$ for ablation study.
\end{itemize}
Additionally, since our algorithm is based on MASAC, we combine COMA, QMIX, and COM with MASAC to obtain the other three baselines: COMA+SAC, QMIX+SAC, and COM+SAC. Notice that the method in \cite{DBLP:journals/corr/abs-1902-03987} is a model-based version of COM and other baselines (including ours) are model-free methods. Thus, we only compare to COM considering fairness.

\begin{figure}[!t]
\setlength{\belowcaptionskip}{-0.3cm}
\setlength{\abovecaptionskip}{0cm}
 \centering
 \subfigure[Precision]{
 \includegraphics[width=.45\textwidth]{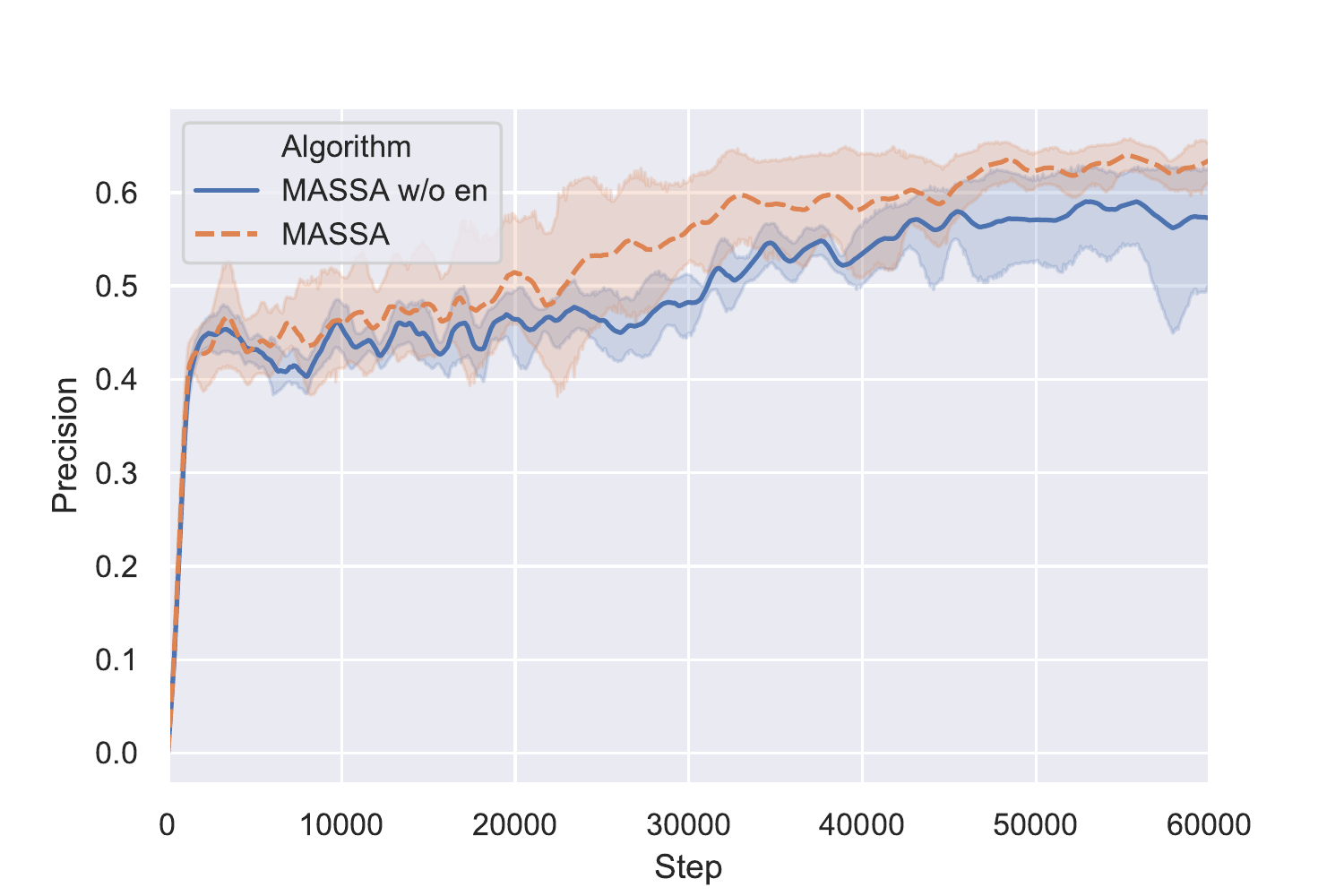}
 }
 \subfigure[nDCG]{
 \includegraphics[width=.45\textwidth]{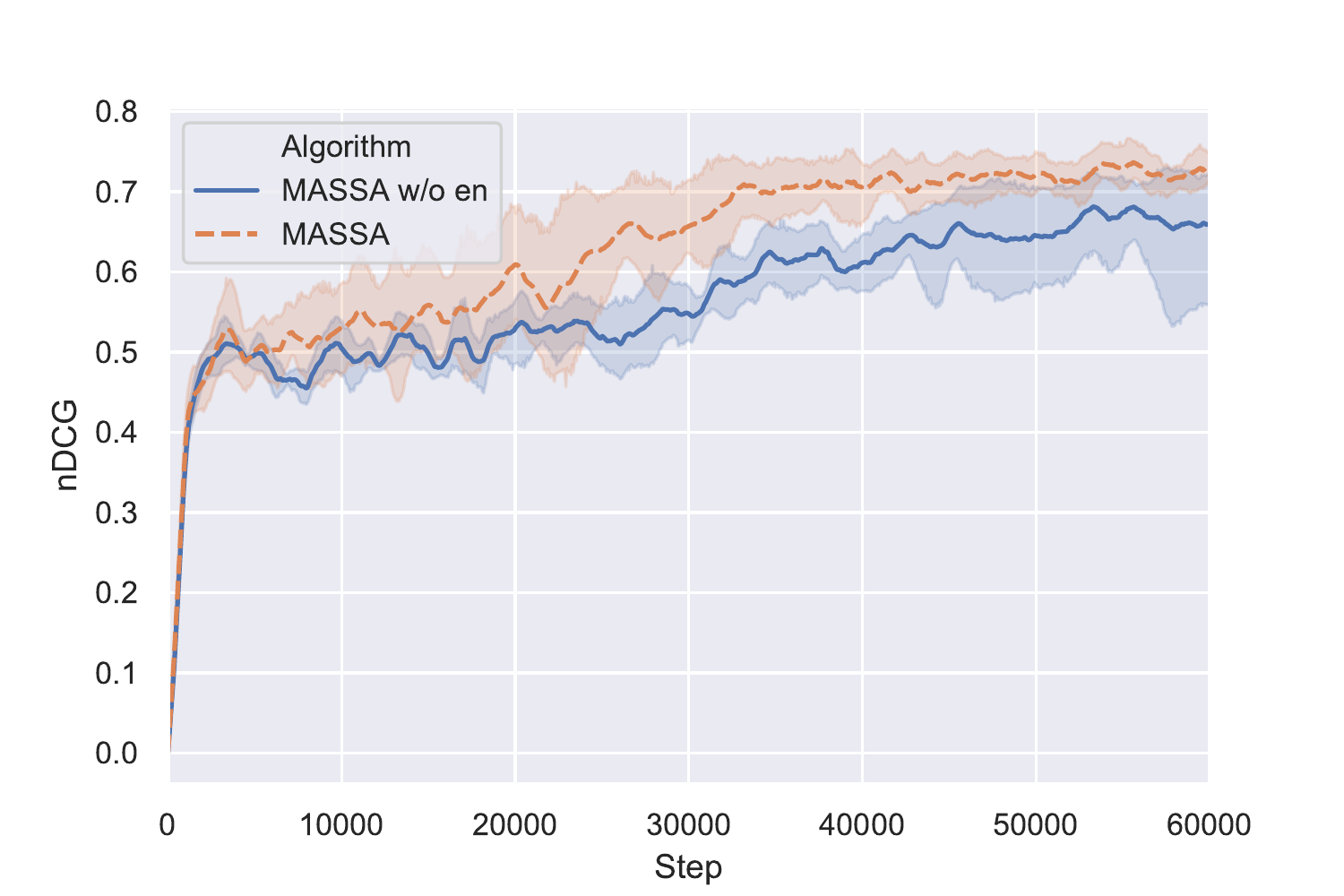}
 }
 \caption{The curves of precision and nDCG during online training. The solid curves correspond to the mean and the shaded region to the minimum and maximum values over the 10 runs. The difference between these two algorithms is the entropy term of the loss function for the signal network.}\label{example}
\end{figure}
\begin{figure}[!t]
\setlength{\belowcaptionskip}{-0.3cm}
\setlength{\abovecaptionskip}{0cm}
 \centering
 \subfigure[Offline]{
 \includegraphics[width=.45\textwidth]{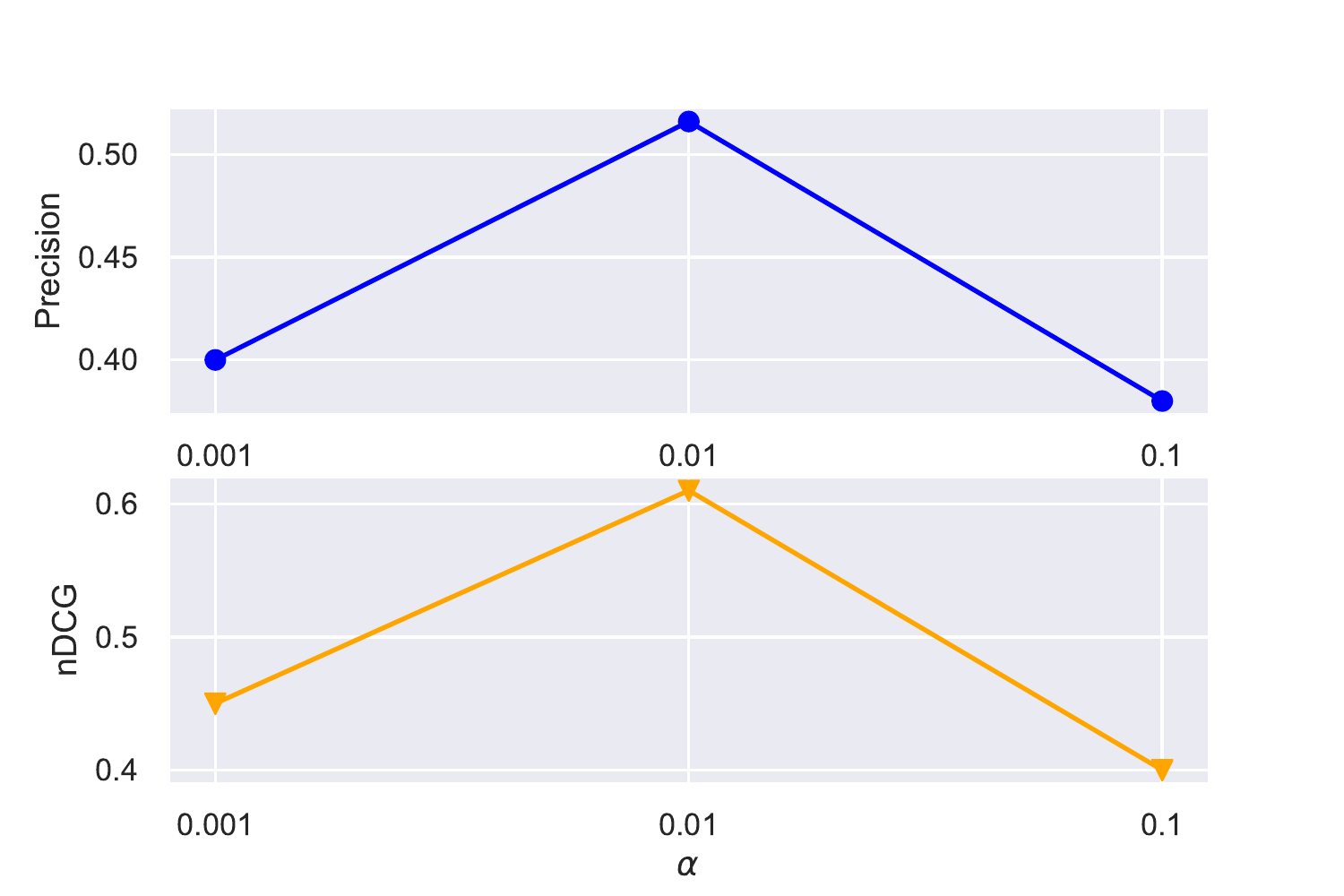}
 }
 \subfigure[Online]{
 \includegraphics[width=.45\textwidth]{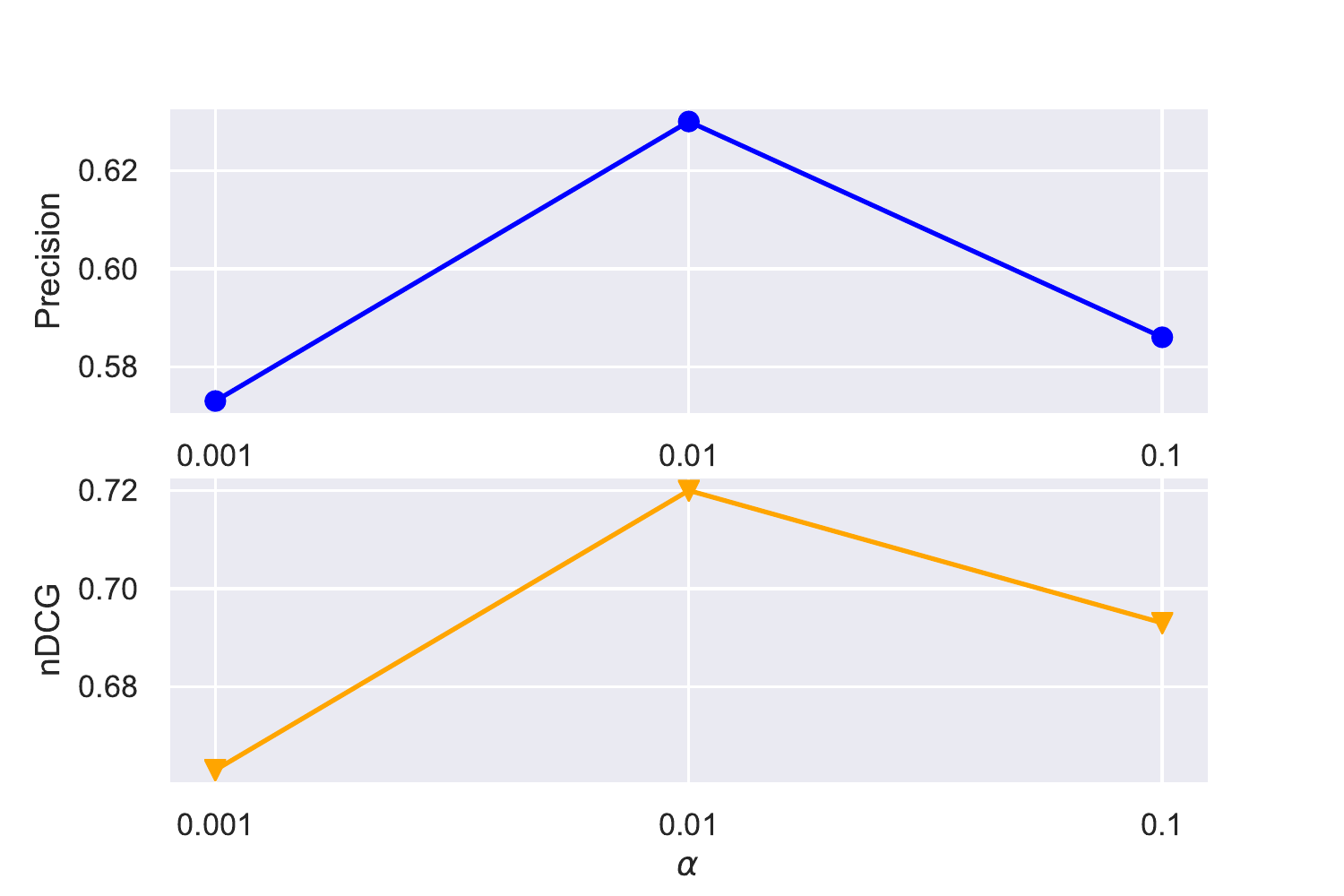}
 }
 \caption{The performance with the change of $\alpha$}\label{alpha}
\end{figure}

\subsection{Result}
In this subsection, we illustrate performance of different methods to indicate the improvement caused by the signal network and the entropy scheme.
\subsubsection{Offline Testing}
The results of our offline evaluation are shown in Table \ref{offline_result}. All the methods are trained by 14-day training data and tested by the 3-day testing data. There are a few interesting conclusions drawn from the results. 

Firstly, the signal network and additional entropy terms can improve performance significantly. Since the only distinction between \textit{MASAC} and \textit{MASSA w/o entropy} is the signal network, the gap of the performance shows the effectiveness of the signal network. Besides, by adding the entropy term to encourage exploration, \textit{MASSA} method outperforms all the other methods. Comparing to \textit{MASSA}, the \textit{MASSA w/o entropy} method is prone to converge to a sub-optimal policy in our scenarios. The entropy-regularized algorithm can explore in view of global performance.

Secondly, the metrics of module 1 are better than that of module 2, which is caused by different properties of these two modules. As shown in Fig. \ref{flow}, module 1 is at the top of the web page. Thus, users are more likely to be attracted by module 1 and ignore another module, especially when the items recommended by module 1 are good. In our dataset, the ratio of the number of clicks in these two modules is about 6:1. 

% Thirdly, the metrics of the two modules increase simultaneously by applying cooperative methods. Notice that if the overall performance of a method $A$ is better than another one $B$, method $A$ almost outperforms method $B$ in view of individual modules. With a cooperative mechanism, the competition of two modules is mitigated and the cooperation leads to a win-win result. Different cooperative methods decide the effect of cooperation. 

Finally, \textit{DDPG} performs worse comparing with \textit{MADDPG} whose actors have a similar structure with  \textit{DDPG}. The main reason is that the ranking policies of the two modules are trained individually without any cooperation. 

\subsubsection{Online Testing}
For the online experiment, Fig. \ref{result_online} exhibits the performance of various algorithms. The performance is the mean of 10 runs. Our algorithms outperform others again in the online experiment.

Firstly, the performance of the methods based on \textit{MASAC} is better than that based on \textit{MADDPG} except for \textit{COMA}. The reason is that the online environment is more complex than the offline setting in terms of the number of candidate items and the source of clicks. The number of candidate items increases from 10 to 2000 for each set and the clicks are from an online simulator. Exploration is more important to obtain a better policy in a complex environment. Thus, SAC-based approaches perform better.

Secondly, although \textit{MASSA w/o entropy} is better than \textit{MASSA} for the module 2, the overall performance of \textit{MASSA w/o entropy} is worse. It illustrates that in order to find a globally optimal solution, \textit{MASSA} makes a small sacrifice of module 2 and obtains a huge improvement for the overall performance. 

\textbf{The effect of entropy}
The importance of the entropy term is indicated in Fig. \ref{example}. We can observe that two algorithms perform similar in the first 20 thousand steps and the overall performance seems to be constant if we ignore the perturbation, which means that the ranking policy falls into a sub-optimal solution. Due to the entropy term, \textit{MASSA} constantly explores and escapes from the sub-optimal solution at around 30 thousand steps. Finally, a globally optimal solution is found. However, \textit{MASSA w/o entropy} algorithm only finds a better sub-optimal solution slowly.

Another interesting fact is the change in the shaded region. For \textit{MASSA}, the region is huge before 45 thousand steps and becomes smaller in the last 10 thousand steps. However, the region of \textit{MASSA w/o entropy} becomes larger at the end of the training. It indicates that \textit{MASSA} explores more at the beginning and converges to the optimal solution. However, due to the lack of exploration, \textit{MASSA w/o entropy} falls into different sub-optimal solutions in the end.

% \textbf{examples of actions}
% \begin{figure}[!t]
%   \centering
%   \includegraphics[width=.3\textwidth]{alpha.pdf}\\
%   \caption{The performances with the change of $\alpha$}\label{alpha}
% \end{figure}

\subsubsection{The influence of $\alpha$}
Fig. \ref{alpha} shows the performance with the change of $\alpha$ which is the weight of the entropy term for the loss function of the signal network. Our algorithm performs the best when $\alpha=0.01$. Thus, we use this value in both online and offline experiments. 

\section{Conclusion}
In this paper, we propose a novel multi-agent cooperative learning algorithm for the multi-module recommendation problem, in which a page contains multiple modules that recommend items processing different specific properties. To prompt cooperation and maximize the overall reward, we firstly design a signal network that sends additional signals to all the modules. Secondly, an entropy-regularized version of the signal network is proposed to coordinate agents' exploration. Finally, we conduct both offline and online experiments to verify that our proposed algorithm outperforms other state-of-the-art learning algorithms.

\begin{acks}
This work was supported by Alibaba Group through Alibaba Innovative Research (AIR) Program and Alibaba-NTU Joint Research Institute (JRI), Nanyang Technological University, Singapore.
\end{acks}

\bibliographystyle{ACM-Reference-Format}
\bibliography{acmart}

% \appendix

\end{document}